\documentclass[11pt]{article}
\usepackage{algorithm2e}
\usepackage[dvipsnames]{xcolor}
\usepackage{amssymb}
\usepackage{authblk}
\usepackage[LFE,LAE,T1]{fontenc}
\usepackage{amsmath}
\usepackage{comment}
\usepackage{graphicx}
\usepackage{blindtext}
\usepackage{ragged2e}
\usepackage{subcaption}
\usepackage{setspace}

\usepackage{wrapfig}
\usepackage{fancyhdr}
\usepackage{cite}
\usepackage[a4paper, total={17cm, 26.7cm}]{geometry}
\usepackage{times}

\title{Generative AI for Hate Speech Detection: Evaluation and Findings}
\author[1]{Sagi Pendzel}
\author[2]{Tomer Wullach}
\author[3]{Amir Adler}
\author[1]{Einat Minkov}
\affil[1]{University of Haifa}
\affil[2]{OriginAI}
\affil[3]{Braude College of Engineering}

\begin{document}
\maketitle
\doublespacing
\begin{abstract}
Automatic hate speech detection using deep neural models is hampered by the scarcity of labeled datasets, leading to poor generalization. To mitigate this problem, generative AI has been utilized to generate large amounts of synthetic hate speech sequences from available labeled examples, leveraging the generated data in finetuning large pre-trained language models (LLMs). In this chapter, we provide a review of relevant methods, experimental setups and evaluation of this approach. In addition to general LLMs, such as BERT, RoBERTa and ALBERT, we apply and evaluate the impact of train set augmentation with generated data using LLMs that have been already adapted for hate detection, including RoBERTa-Toxicity, HateBERT, HateXplain, ToxDect, and ToxiGen. An empirical study corroborates our previous findings, showing that this approach improves hate speech generalization, boosting recall performance across data distributions. In addition, we explore and compare the performance of the finetuned LLMs with zero-shot hate detection using a GPT-3.5 model. Our results demonstrate that while better generalization is achieved using the GPT-3.5 model, it achieves mediocre recall and low precision on most datasets. It is an open question whether the sensitivity of models such as GPT-3.5, and onward, can be improved using similar techniques of text generation.
\end{abstract}
\renewcommand{\headrulewidth}{0pt} 

\centering

\normalsize

\justifying

\section{Introduction}

{\it Hate speech} refers to the expression of hateful or violent attitudes based on group affiliation such as race, nationality, religion, or sexual orientation. In light of the increasing prevalence of hate speech on social media, there is a pressing need to develop automatic methods that detect hate speech manifestation at scale~\cite{fortuna18}. Automatic methods of natural language processing in general, and hate speech detection in particular, heavily rely on relevant datasets. While researchers have collected several datasets that contain hate speech samples, those resource are scarce. Furthermore, the difficulty in identifying hate speech on social media has led to the use of biased data sampling techniques, focusing on a specific subset of hateful terms or accounts. Consequently, relevant available datasets are limited in size, highly imbalanced, and exhibit topical and lexical biases. Several recent works have indicated these shortcomings, and shown that classification models trained on those datasets merely memorize keywords, where this results in poor generalization~\cite{wiegandNAACL19,kennedyACL20}.

In this chapter, we describe a recent line of research which seeks to improve hate speech detection via the synthesis of large corpora of hate speech utterances~\cite{wullach21}. Indeed, it has been shown that augmenting labeled datasets with synthetic text sequences improves the generalization of hate speech detection using state-of-the-art pretrained large language models (LLMs)~\cite{wullachEMNLP21}. In general, we employ transformer-based language encoders like BERT~\cite{devlin2018bert}, RoBERTa~\cite{roberta19}, and their variants as hate speech classifiers. These encoders have been pretrained on massive heterogeneous text corpora with the objective of encoding text semantics within a low-dimensional vector space. In order to perform text classification, the network parameters of the pretrained models are typically adapted to a target task using supervised training via a {\it finetuning} procedure~\cite{devlin2018bert}. Due to the deep language representations encoded in these LLMs, they typically achieve improved performance in low-resource classification settings~\cite{kennedyACL20}. Yet, large volumes of high-quality labeled examples must be provided to achieve high model generalization~\cite{gururanganACL20}. 

In several recent works, we described a method for extending available manually-curated hate speech datasets with large amounts of generated labeled examples. We generated a large number of synthetic text sequences using the LLM decoder of GPT-2~\cite{gpt2}, having it tuned using human-labeled examples to generate hate (and non-hate) speech ~\cite{wullach21}. We then augmented the existing gold-labeled datasets with large amounts of synthetic examples, increasing their size by several magnitudes of order. In experiments with pre-trained language models like BERT, RoBERTa, and ALBERT~\cite{albertICLR2020}, we observed substantial and consistent improvements when using synthetic data. Notably, we showed improved generalization in cross-dataset evaluation, simulating the realistic scenario where there is a distribution shift between the data that the model is trained and tested on. 

In this chapter, we assess several additional, recently proposed, LLMs, which have been specialized on the task of hate speech detection using additional amounts of relevant data. We examine whether incorporating large volumes of synthetic hate speech examples in further finetuning these models improve their generalization in the realistic cross-dataset evaluation setting. Our results show that this is indeed the case. In accordance with our previous findings~\cite{wullachEMNLP21}, we find that finetuning the models using large amounts of synthetic examples often leads to dramatic improvements in recall. In another experiment, we consider Toxigen, another corpus of synthetic hate speech sequences, which is targeted at representing implicit hate speech statements. We show that a mixture of the two corpora yields the best improvements in recall, probably due to increased data diversity. Finally, we consider GPT-3.5 as a model of hate speech detection. While it is a proprietary model, GPT-3.5 is admittedly a very large language model, which has been trained using vast amounts of labeled and unlabeled data, and has been tuned to process and generate text given human feedback with respect to multiple goals, one of which is presumably toxicity detection~\cite{ye2023comprehensive}. Evaluating GPT-3.5 on our test sets reveals that it is recall-oriented, and yields the best recall as well as F1 (the harmonic mean of precision and recall) across all of the evaluated methods. Nevertheless, the performance of all methods is roughly `at the same ballpark', reaching similar levels of recall. We attribute this to our dataset augmentation procedure. 

Importantly, we view that the challenge of automatic hate speech detection is far from being solved--the best models often fail to recognize some hate speech utterances, as reflected by lower-than-desired recall rates. And, while state-of-the-art models are sensitive to toxicity, prioritizing recall, precision is substantially lower. An open question of interest is whether and how data augmentation via speech synthesis can further enhance high-performing models like GPT-3.5 on the task of hate speech detection.

The rest of this chapter is organized as follows: Section~\ref{sec:related} describes related work on LLMs adapted to hate speech detection, hate speech generation and other related research directions. Section ~\ref{sec:methods} describes our recently proposed method for synthetic hate speech generation~\cite{wullach21}. The experimental setups for assessing hate speech detection using generated hate speech data are presented in section~\ref{sec:experiments}, and the results are detailed in section \ref{sec:results}. Section~\ref{sec:chatgpt} presents the application of GPT-3.5 as a hate detector, along with performance evaluation, followed by concluding remarks in section~\ref{sec:conclusion}.

\section{Related work}
\label{sec:related}

This section first reviews recent content-based models which adapt large pre-trained LMs to the task of hate speech detection using large amounts of related data. Our focus in this chapter is on text generation as means of enriching the labeled data that is provided to such models. Accordingly, we then describe research efforts that use text generation for hate speech detection. Aiming to provide a broader view on the task of hate speech detection, we also refer the reader to several related works that combine semantic encodings of text with network information. This research direction is complementary to our work, and we believe that it holds promise for further improving hate and toxicity detection in social networks. 

\subsection{LLMs specialized for hate speech detection}

Large language models (LLMs) are extensive transformer-based architectures, which were trained to transform text into contextual semantic encodings given vast amounts of unlabeled text. LLMs may be further specialized to a particular domain and task by continued pre-training using related data, or via finetuning on labeled task-related examples~\cite{gururanganACL20}. It is non-trivial however to identify, obtain and employ relevant data. Below, we briefly describe several recent models, which enhance LLMs with relevant data either using extended pre-training or finetuning with the purpose of improving hate speech detection. In our experiments, we examine the impact of further finetuning these specialized models using large amounts of synthetic examples. 

{\it HateBERT}~\cite{hatebert} is a specialized version of the BERT-base model~\cite{bert}, which has been adapted to detect abusive language via pre-training. Concretely, extended pre-training was performed using more than a million posts from Reddit communities, which have been banned due to offensive, abusive, or hateful content. {\it HateXplain}~\cite{mathew2020hatexplain} is another variant of the BERT-base model, which has been specialized to the tasks of hate speech detection and explanation via finetuning using a dedicated benchmark dataset. The HateXplain dataset includes about 20K posts, sampled from Twitter and Gab\footnote{https://gab.com/}, which are annotated with respect to multiple perspectives: whether a post is hateful, offensive, or normal; the target community that is victimized in the post; and, the rationales or the specific portions of the post upon which the labeling decision is based. ToxDect~\cite{toxdect} is another hate detection model, which employs the RoBERTa-large model, a larger and more elaboratively pre-trained LLM, which is generally better-performaing than BERT~\cite{roberta19}. ToxDect utilized the Founta dataset~\cite{fortuna18} for finetuning purposes, having the hateful and abusive classes merged into a single `toxic' class. In order to mitigate biases in toxic language detection, an ensemble learning approach was employed, with the goal of identifying and reducing the importance of biased features in the final model. In general, the less bias a model is, the better it is expected to perform across different data distributions. Finally, the {\it Toxigen}~\cite{toxigen} model is the outcome of finetuning the specialized HateBERT model on large amounts of synthetic examples. Similar to our approach, the authors used a GPT model to generate hateful and non-hateful text sequences, resulting in the Toxigen dataset~\ref{sec:text_generation}. Unlike our work, the Toxigen dataset was designed with the aim of representing implicit hate speech, mimicking tones and styles of hate speech directed at a large variety of social groups. In our experiments, we consider the finetuned Toxigen model in its final form as a task-specialized LLMs, as well as evaluate the Toxigen dataset as an alternative and complementary resource of synthetic examples. A more detailed description of their text generation approach follows.

\subsection{Text generation for hate speech detection}
\label{sec:text_generation}

Human labeled datasets are generally of high relevance and quality, yet they are small and biased with respect to data distribution, where these fallbacks hamper learning generalization~\cite{wiegandNAACL19}. Researchers have therefore attempted various ways of extending labeled datasets using automatic approaches. One manner in which new sequences can be obtained is {\it back-translation}. Using this approach, labeled examples are automatically translated to another language, and are then back-translated to the source language. This method has been widely used to enhance text translation models, as it generates many aligned sentence pairs, which presumably preserve the original sentence meaning, while introducing lexical variance~\cite{beddiar2021data}. Yet, the capacity of this method in generating new examples is limited. Instead, we opt for generating new text sequences, which divert from the original dataset. Given labeled examples, we tuned a generative LLM, specifically, a GPT-2 model, to synthesise class-dependent text sequences that are either hateful or non-hateful. In comparison to back-translation, this approach is not constrained to closely preserve the original texts, and is therefore more scalable. In our work, we generated a corpus of synthetic text that is three times of order larger than the original datasets, reaching 2M sequences. While text generation increases data diversity, it introduces some noise, either with respect to semantic relevance or syntax. Consequently, we observed that the augmentation of human labeled datasets with synthetic examples results in substantial gains in recall, while precision is impaired~\cite{wullach21,wullachEMNLP21}.  Considering that hate speech is a minority class, and that there are severe consequences of failing to identify hate speech, we argue that utmost importance should be attributed to achieving high recall. 

A few previous works used synthetic text sequences for related purposes. Similar to our approach,~\cite{tavorAAAI20} synthesized new examples from existing training data. They finetuned GPT-2 by prepending the class label to text samples, and then generated new sentences conditioned on the class label. They rather focused on balancing multi-class datasets however, generating up to several thousands of examples per class~\cite{tepperEMNLP20}. More related to our task,~\cite{toxigen} generated synthetic hateful and neutral text sequences. Aiming at representing implicit hate speech, they used crowd sourcing to obtain implicit hate speech utterances, as well as neutral statements. The elicited examples were required to mention a variety of minority identity groups, in order to mitigate lexical biases in the generated examples. Employing GPT-3, they created a dataset of synthetic examples, named {\it ToxiGen}, which includes 274K synthetic examples, balanced across toxic and benign statements targeted at multiple minority groups. In our experiments, we show that also when using this resource for dataset augmentation, recall typically improves, supporting our general claims about the efficacy of employing data synthesis for finetuning LLMs.

A recent study~\cite{casula2023generation} which explored offensive language detection using generation-based data augmentation, suggested that while this approach can occasionally enhance model performance, its impact is inconsistent. In addition, they found that generative data augmentation can introduce unpredictable lexical biases. The experimental results that we report in this chapter are in line with their findings. While we do not study lexical biases, our experiments demonstrate high variance with respect to the impact of data augmentation across datasets and methods. Nevertheless, we do show that there are consistent and substantial improvements achieved in terms of recall using this approach.

A couple of other related works which applied text generation in the context of hate speech detection include another previous work, where we demonstrated dramatic improvement of small hate speech classifiers, targeted at end devices of limited computation capacity, following training using synthetic examples~\cite{wullachESWA22}. In another work of interest, researchers proposed to employ LLMs for the automatic generation of counter hate speech~\cite{counter2022}. They applied prompt engineering to generate counter speech using models such as GPT-2 and GPT-3, indicating that this approach forms a promising direction for combating hate speech online. 

\subsection{Contextual hate speech detection}

In this chapter, we focus on content-based approaches to hate speech detection. This approach is arguably inherently limited in that the texts posted on social media are inherently short, and lack sufficient context information. The modeling of relevant context is crucial when rhetoric elements such as sarcasm are used. Researchers have indicated on various contextual evidence that can be used in determining text toxicity~\cite{gao2017}. One may refer to additional texts about the topic discussed, or previous postings within the same thread, e.g.,~\cite{perez2023assessing}. In addition, information about the author of the post, based on their previous posts or network information~\cite{lotanPLOS23}, may serve as meaningful evidence in inferring the text meaning as intended by them~\cite{chakraborty22,lotanPLOS23}. Our work is orthogonal to those efforts. Once LLMs are tuned to produce improved semantic encodings of the text for the task at hand, it is possible to integrate these encodings with additional evidence types using dedicated classifiers~\cite{penzel2023detecting}. We believe this to be a promising direction of future research.

\section{Synthetic hate speech generation: Method and the MegaSpeech corpus}
\label{sec:methods}

In our work, we proposed to
exploit existing gold-labeled datasets, which are limited in size, for the generation of large amounts of related, pseudo-labeled, synthetic text sequences~\cite{wullach21}. More formally, given a dataset $d^i$ that consists of hate and non-hate labeled examples $\{d^i_h,d^i_{nh}\}$, it is desired to generate additional class-conditioned synthetic text sequences. While various generative language models may be used, our work utilized the model of GPT-2 (764M parameters). Notably, larger and improved generative models exist as of today, which have shown to produce yet better text sequences, of improved semantic and syntactic quality~\cite{ye2023comprehensive}. The application of the framework to those models is a direction of future work.

The approach is based on the following principles. 
\begin{enumerate}
\item In order to bias the model towards the genre of micro-posts, hate speech, and the topics and terms that characterise each dataset, we continue training the GPT model from its distribution checkpoint, serving it with the labeled text sequences. Concretely, we adapt distinct models per dataset and class, i.e., for each dataset $d^i$, we obtain two models, $G^i_h$ (hate speech) and $G^i_{nh}$ (non-hate). 
\item In text synthesis, we provide no prompt to the respective GPT model, that is, the token sequences are generated unconditionally, starting from the empty string. Similar to the labeled datasets, we generate sequences that are relatively short, up to 30 tokens. 
\item Presumably, not all of the text sequences generated by $G^i_h$ are hateful. We utilize the labeled examples $d^i$ for finetuning a classifier (BERT) on hate detection, and apply the resulting classifier to the sequences generated by $G^i_h$. Only those sequences that are perceived as hateful by the model, for which the prediction confidence scores are high, are maintained. In our experiments, we set the threshold to 0.7, discarding about two thirds of the generated hate speech sequences. 
\item Finally, we augment the labeled examples $d^i$ with an equal number of hate and non-hate synthetic examples. 
\end{enumerate}

\begin{figure}[t]
\centering
\includegraphics[trim=0.02cm 0.02cm 0.02cm 0.96cm, clip,width=0.95\textwidth]{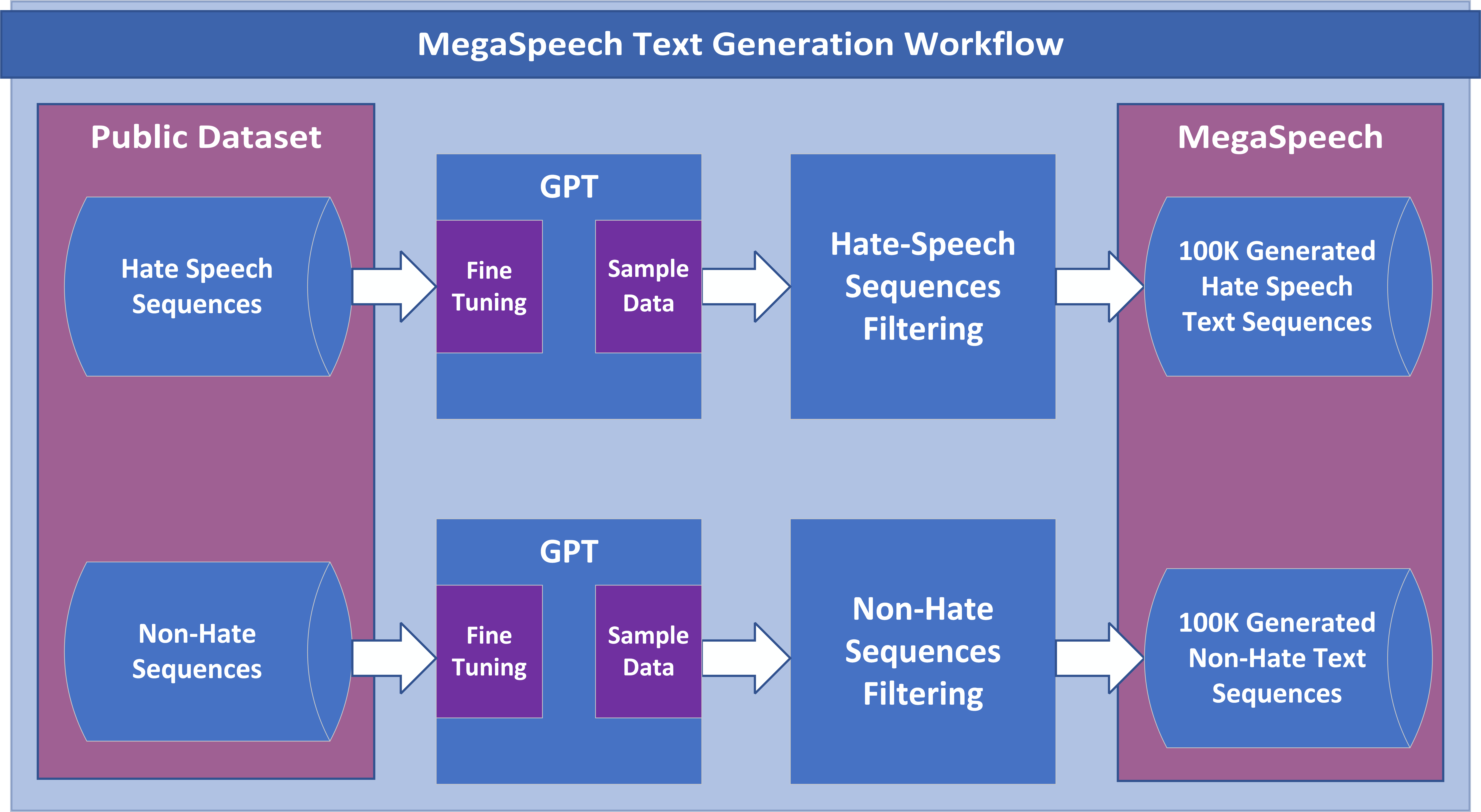}
\captionsetup{font=small,labelfont={bf,sf}}
    \caption{The text generation workflow, performed using the method described in Sec.~\ref{sec:methods} and each one of the five baseline public datasets in Table \ref{tab:datasets} (200K per dataset), resulted in a total of 1M generated text sequences, which comprise the MegaSpeech corpus\cite{wullach21}.}
    \label{fig:megaspeech}
\end{figure}

\paragraph{The MegaSpeech corpus}
Applying this procedure to the datasets described in Table~\ref{tab:datasets}, we created a large resource of synthetic hateful and non-hateful examples. Specifically, we generated 200K text sequences per dataset: 100k per the hate speech class and 100k per the neutral class. Importantly, a subset (20\%) of each dataset was excluded for both text generation and evaluation purposes, comprising out test sets. The resulting corpus, named MegaSpeech, includes 1M sequences overall. Figure~\ref{fig:megaspeech} provides an illustration of the MegaSpeech corpus generation process. 

\section{Experiments}
\label{sec:experiments}

\begin{table}[t]
\centering
\begin{tabular}{llrc}
     Dataset & Source & Size [K] & Hate class ratio\\
\hline
DV & \cite{davidson2017automated} &  {6} & {0.24}\\
FT & \cite{founta2018large} &  {53} & {0.11}\\ 
WS & \cite{waseem2016hateful} & {13} & {0.15}\\
SF (StormFront) & \cite{de2018hate} &  {9.6} & {0.11}\\
SE (SemEval) & \cite{basile2019semeval} & {10} & {0.40}\\
\hline
\end{tabular}
\captionsetup{font=small,labelfont={bf,sf}}
\caption{The experimental hate speech datasets}
\label{tab:datasets}
\end{table}

We wish to assess whether and to what extent the generated synthetic data is sufficiently relevant and diverse for improving the generalization of pretrained LMs on the hate detection task. Compared with our previous work~\cite{wullachEMNLP21}, our current experiments include multiple state-of-the-art LLMs which were specialized for hate speech detection, where we show that the synthetic examples can be used to enhance them further. Our evaluation is focused on a cross-dataset setup, testing the generalization of the models across data distributions. Throughout this work, performance is reported in terms of precision, recall and F1 with respect to the hate class. Considering that hate speech is a minority class within the general data stream in social media, and that the harm caused by hate speech is high, we place emphasis on recall~\cite{wullach21}.

\subsection{Datasets}

Table~\ref{tab:datasets} provides details about the experimental datasets. Some of the datasets originally used a fine annotation scheme, e.g., distinguishing between hate speech and abusive language. Since we perform transfer learning across datasets, we maintain the examples strictly annotated as hate and non-hate, and discard the examples assigned to other categories. 

As shown, the datasets are small (6-53K labeled examples) and skewed, with as little as 1-6k hate speech examples available per dataset. All of the datasets include tweets, except for SF, which includes individual sentences extracted from the StormFront Web domain. It has been previously shown that these datasets exhibit various biases, stemming from the underlying data collection procedure~\cite{wiegandNAACL19}, where this prohibits generalization. Additional details about these datasets, as well as examples of the tweets generated per dataset, are available in ~\cite{wullach21}. All datasets were randomly split into train (80\%) and test (20\%) sets, while maintaining similar class proportions. Only the train examples were used in the sequence generation process.

\subsection{Experimental setup}

\paragraph{Models.} In our previous experiments~\cite{wullachEMNLP21}, the popular RoBERTa model~\cite{roberta19} yielded comparable or preferable hate detection results compared with the models of BERT and ALBERT~\cite{albertICLR2020}. Here, we consider {\it RoBERTA-Toxicity}, a variant of RoBERTa which has been finetuned on the Jigsaw toxic online comment classification datasets, encompassing approximately 2M toxic and benign comments.\footnote{Concretely, the model was finetuned using the English parts of multiple Jigsaw datasets: Jigsaw 2018, https://www.kaggle.com/c/jigsaw-toxic-comment-classification-challenge; Jigsaw 2019, https://www.kaggle.com/c/jigsaw-unintended-bias-in-toxicity-classification; and Jigsaw 2020, https://www.kaggle.com/c/jigsaw-multilingual-toxic-comment-classification.} In preliminary experiments, this model was yielded comparable and sometimes better results than the benchmark version of RoBERTa. In the experiments, we further evaluate the performance of specialized hate speech LLMs described in Section~\ref{sec:related}.
All models were applied using their public implementations, which are available on the HuggingFace platform.\footnote{RoBERTa-Toxicity: https://huggingface.co/s-nlp/roberta\_toxicity\_classifier/; HateBERT: https://huggingface.co/GroNLP/hateBERT; ToxDect: https://huggingface.co/Xuhui/ToxDect-roberta-large; Toxigen: https://huggingface.co/tomh/toxigen\_hatebert; HateXplain: https://huggingface.co/Hate-speech-CNERG/bert-base-uncased-hatexplain}

\paragraph{A cross-dataset evaluation setup.}

In applying hate speech detection models, it is likely that the target distribution of hate speech differ or vary over time from the train set distribution. A realistic evaluation of hate speech detection model must therefore test their generalization in conditions of transfer learning, having the models trained and tested using examples drawn from different datasets~\cite{wiegandNAACL19}. 

As expected, we observed steep degradation in performance of LLMs trained and tested across dataset pairs. We further showed that augmenting the labeled datasets with a large number of synthetic examples improves cross-dataset generalization~\cite{wullach21}. Along these lines, we consider here a cross-dataset evaluation setup. Similar to our previous work~\cite{wullachEMNLP21}, we opt for a resource-inclusive cross-dataset learning and evaluation strategy, where we finetune the various models using the labeled examples of multiple (4) datasets, and then apply the adapted models to predict the labels of the remaining held-out dataset. As shown by us and other researchers~\cite{antypasWS23}, this training strategy yields better generalization compared with a procedure that uses some homogeneous dataset in training.

Considering the cross-dataset evaluation setup as proxy to hate speech detection `in the wild', we wish to gauge the potential benefit of using synthetic examples for learning more effective models. Accordingly, we experiments and report our results for the following experiment sets:
\begin{itemize}
\item {\it 4-vs-1.} Provided with 5 datasets~\ref{tab:datasets}, we perform and report five experiments. In each experiment, the specified LLM is trained using the labeled examples of 4 datasets, and tested on the labeled examples of the remaining held-out dataset.
\item {\it 4-vs-1: Gen} For each one of the experiments above, we finetune the same LLM using the same labeled examples, incorporating additional 240K synthetic examples. The added examples were randomly selected from the MegaSpeech corpus. As described in Section~\ref{sec:methods}, each synthetic example was generated so as to match the language of a given dataset. In selecting the random example, we maintain a balance across source dataset and class. The models finetuned using the augmented datasets are tested on the labeled examples of the held-out dataset, allowing a direct comparison with the non-augmented finetuning experiments. 
\end{itemize}

\paragraph{Implementation details.}

In both sets of experiments, we split the training examples into stratified train (90\%) and validation (10\%) sets, fine-tuning the parameters to values that optimize the cross-entropy loss on the validation examples. Each experiment was conducted up a maximum of 3 training epochs, randomly shuffling the training examples. Evaluations were conducted at intervals of 0.25 epoch, employing an early stopping mechanism. In practice, the majority of experiments terminated after 0.75 epochs. A mini-batch size of 32 was employed in combination with the Adam optimizer, initialized with a learning rate of 2e-5 and 200 warm-up steps. The experiments were conducted using a NVIDIA Tesla P100 GPU and 16GB RAM as the implementation environment.

\section{Experimental Results}
\label{sec:results}
\subsection{Main findings}

\begin{table}[p]
\centering
\begin{tabular}{l|ccc|ccc}
    & \multicolumn{3}{c|}{4-vs-1} &  \multicolumn{3}{c}{4-vs-1: Gen [240K]} \\
    \hline
    & Precision & Recall & F1 & Precision & Recall & F1 \\
\hline
\multicolumn{7}{l}{\textbf{FT}} \\
\hline
RoBERTa-Tox.	&	\textbf{0.673}	&	0.424	&	\textbf{0.521}	&	0.397	&	\textbf{0.569}	&	0.467	\\
ToxiGen	&	0.313	&	0.325	&	0.319	&	0.439	&	0.421	&	0.430	\\
HateXplain	&	0.561	&	0.358	&	0.437	&	0.471	&	0.500	&	0.485	\\
ToxDect	&	0.552	&	0.473	&	0.509	&	0.514	&	0.445	&	0.477	\\
HateBERT	&	0.508	&	0.464	&	0.485	&	0.373	&	0.536	&	0.440	\\
\hline									
\multicolumn{7}{l}{\textbf{SF}} \\			\hline				
RoBERTa-Tox.	&	0.597	&	0.627	&	0.612	&	0.468	&	0.742	&	0.574	\\
ToxiGen	&	0.525	&	0.456	&	0.488	&	0.449	&	0.632	&	0.525	\\
HateXplain	&	0.559	&	0.680	&	\textbf{0.614}	&	0.470	&	0.739	&	0.574	\\
ToxDect	&	\textbf{0.599}	&	0.566	&	0.582	&	0.388	&	0.781	&	0.519	\\
HateBERT	&	0.479	&	0.792	&	0.597	&	0.396	&	\textbf{0.797}	&	0.529	\\
\hline
\multicolumn{7}{l}{\textbf{DV}} \\	
\hline
RoBERTa-Tox.	&	0.762	&	0.820	&	\textbf{0.790}	&	0.692	&	0.691	&	0.692	\\
ToxiGen	&	\textbf{0.867}	&	0.369	&	0.517	&	0.698	&	0.557	&	0.620	\\
HateXplain	&	0.721	&	0.720	&	0.720	&	0.633	&	0.663	&	0.648	\\
ToxDect	&	0.647	&	0.835	&	0.729	&	0.631	&	0.741	&	0.682	\\
HateBERT	&	0.713	&	\textbf{0.853}	&	0.776	&	0.680	&	0.764	&	0.719	\\
\hline
\multicolumn{7}{l}{\textbf{SE}} \\									\hline
RoBERTa-Tox.	&	0.642	&	0.472	&	0.544	&	0.539	&	0.743	&	0.625	\\
ToxiGen	&	\textbf{0.665}	&	0.319	&	0.432	&	0.545	&	0.633	&	0.585	\\
HateXplain	&	0.659	&	0.446	&	0.532	&	0.555	&	0.688	&	0.614	\\
ToxDect	&	0.576	&	0.678	&	0.623	&	0.517	&	0.739	&	0.608	\\
HateBERT	&	0.622	&	0.657	&	\textbf{0.639}	&	0.539	&	\textbf{0.764}	&	0.632	\\
\hline
\multicolumn{7}{l}{\textbf{WS}} \\									\hline		
RoBERTa-Tox.	&	0.861	&	0.673	&	0.756	&	0.853	&	0.856	&	0.854	\\
ToxiGen	&	0.770	&	0.478	&	0.590	&	0.881	&	0.790	&	0.833	\\
HateXplain	&	\textbf{0.891}	&	0.659	&	0.758	&	0.848	&	0.865	&	0.856	\\
ToxDect	&	0.625	&	0.844	&	0.719	&	0.831	&	0.902	&	\textbf{0.865}	\\
HateBERT	&	0.791	&	0.891	&	0.838	&	0.753	&	\textbf{0.919}	&	0.828	\\
\hline
\multicolumn{7}{l}{\textbf{Average improvement using the generated examples, per model:}} \\
\hline
RoBERTa-Tox.	&		&		&		&	-17.8\%	&	24.3\%	&	-0.2\%	\\
ToxiGen	&		&		&		&	0.6\%	&	56.5\%	&	27.8\%	\\
HateXplain	&		&		&		&	-13.0\%	&	25.2\%	&	4.6\%	\\
ToxDect	&		&		&		&	-4.3\%	&	7.3\%	&	-1.1\%	\\
HateBERT	&		&		&		&	-13.3\%	&	5.0\%	&	-6.1\%	\\
\hline
\multicolumn{7}{l}{\textbf{Overall average improvement using the generated examples:}} \\
\hline
&		& &		&	-9.6\%	&	23.7\%	&	5.0\%	\\
\hline
\end{tabular}
\captionsetup{font=small,labelfont={bf,sf}}
\caption{Detailed cross-dataset (4-vs-1) results reported in terms of precision, recall and F1 with respect to the hate speech class. The table shows the results of learning models using the original labeled datasets, training the models using 4/5 datasets and evaluating them on the remaining set-aside dataset ('4-vs-1'). The table further shows the results of learning and evaluating the models using the same datasets, having the training datasets augmented with 240K synthetic examples, generated at equal proportions across source dataset and class (`4-vs-1: Gen').} 
\label{tab:cross}
\end{table}

Table~\ref{tab:cross} shows our results without and with train data augmentation, applying the 4-vs-1 cross-dataset experimental setup using the various models. Our examination of the results focuses on the impact of train set augmentation with a large number of synthetic examples on test set performance. In general, we observe mixed trends in the improvement rates across target datasets and methods. Substantial improvements, in all metrics, are observed in most of the experiments when the WhiteStorm dataset is set as the test set distribution. When SemEval is the set-aside dataset, we observe high increase in recall performance, alongside moderate decrease in precision, where the  overall impact on F1 performance is positive. For other datasets, the results differ more significantly depending on the methods employed. Indeed, a related work recently showed that data augmentation results may be inconsistent~\cite{casula2023generation}. Questions regrading the factors that affect learning improvements using synthetic examples in concrete cases remain open.

A higher-level view of the results is provided at the bottom of Table~\ref{tab:cross}, showing a summary of the improvement rates with respect to each metric, averaged across the experiments, per method. This summary illustrates the experimental results with greater clarity. On average, data augmentation leads to improved recall. Recall rates improved by a striking ratio of 24\% or more on average using most of the models. On the other hand, precision decreased in most cases, yet more moderately, reaching a decrease of 17.8\% at the worst case. Accordingly, the average improvement in terms of F1 performance ranged between -6.1\% and 27.8\% across models. A yet broader summary of the results is included in the bottom line of Table~\ref{tab:cross}, averaging the improvement rates with respect to both experiment and model. It is shown that overall, recall improved to a great extent (23.7\%), at the cost of reduced precision (-9.6\%), resulting in overall increase in terms of F1 (5.0\%).

Thus, this set of experiments shows similar trends to the results that we observed in our previous studies~\cite{wullach21,wullachEMNLP21}.
We believe that data augmentation introduces lexical diversity into the training datasets, leading to improved recall in learning from the augmented datasets. On the other hand, the potentially lower quality or relevance of the artificially generated examples could reduce precision. Importantly, we note that the F1 metric attributes equal importance to precision and recall performance. To the extent that detecting a large number of hate speech instances as possible is flagged as top priority, recall is of higher importance in practice. To that end,  dataset augmentation via generation serves to significantly increase recall rates. Importantly, the methods evaluated in this paper have already been specialized on the task of hate speech detection using additional dedicated data. Hence, the reported results corroborate our previous findings, showing that data augmentation can boost recall rates also using task-specialized models. 

\subsection{MegaSpeech vs. ToxiGen: Experiments using different synthetic example distributions}

To assess the impact of text synthesis on hate speech detection performance more broadly, we conducted additional experiments, varying and comparing data augmentation results using different resources of generated examples. Concretely, we consider the ToxiGen corpus as alternative resource of generated hate and non-hate speech examples. As described in Section~\ref{sec:related}, the ToxiGen corpus was generated using the GPT3 model. Unlike the MegaSpeech corpus, which expands existing labeled datasets by means of text generation, ToxiGen was initialized with a relatively small number human-authored texts that  articulate implicit hate speech towards a variety of minority social groups. Hence, the text sequences within ToxiGen are similarly intended to portray implicit hate speech, applying to multiple targets.  

In another set of experiments, we assess hate speech performance using the two different sources of synthetic examples, namely MegaSpeech and Toxigen. For simplicity, we set the learning model of choice to RoBERTa-Toxicity. We maintain the 4-vs-1 dataset evaluation setup. That is, the labeled training and test examples remain the same as in our main experiments. For consistency in training set size, we limit the number of synthetic examples to 240K in all experiments. We drew synthetic examples from the Toxigen corpus randomly and in a stratified fashion, ensuring equal proportions of synthetic examples considered as hateful and non-hateful, as well as similar proportions of text sequences with respect to the various targets, as represented in the Toxigen corpus.

Our experimental results are reported in Table~\ref{tab:toxigen}. For convenience, the table repeats our baseline results as reported in Table~\ref{tab:cross}. These results used the RoBERTa-Toxicity model finetuned with labeled examples (termed `No aug.'). The table further repeats our results using the labeled datasets augmented with synthetic examples from the MegaSpeech corpus (labeled `MegaSpeech'). Alongside those results, produced in our main experiments, the table displays the results using the alternative pool of examples drawn from the ToxiGen corpus for train set augmentation (`ToxiGen'). Finally, we report the results of mixing synthetic examples from the two corpora (`Both'). In this subsequent experiment, we kept the training set size constant, where rater than incorporate 240K examples from a single source, we obtain 120K synthetic examples from each of the MegaSpeech and ToxiGen corpora. The pool of synthetic examples in this setup was selected randomly and in a stratified fashion with respect to example label, origin dataset (MegaSpeech) or target (ToxiGen). 

\begin{table}[p]
\centering
\begin{tabular}{l|ccc}
    \hline
    & Precision & Recall & F1 \\
\hline
\multicolumn{4}{l}{\textbf{FT}} \\
\hline
No augmentation	&	\textbf{0.673}	&	0.424	&	\textbf{0.521}	\\
+ generated examples: MegaSpeech	&	0.397	&	0.569	&	0.467	\\
+ generated examples: ToxiGen	&	0.332	&	0.491	&	0.396	\\
+ generated examples: Mixed	&	0.356	&	\textbf{0.606}	&	0.449	\\
\hline									
\multicolumn{4}{l}{\textbf{SF}} \\			\hline				
No augmentation	&	\textbf{0.597}	&	0.627	&	\textbf{0.612}	\\
+ generated examples: MegaSpeech	&	0.468	&	0.742	&	0.574	\\
+ generated examples: ToxiGen	&	0.179	&	\textbf{0.945}	&	0.301	\\
+ generated examples: Mixed	&	0.185	&	0.936	&	0.309	\\
\hline
\multicolumn{4}{l}{\textbf{DV}} \\	
\hline
No augmentation	&	\textbf{0.762}	&	0.820	&	\textbf{0.79}	\\
+ generated examples: MegaSpeech	&	0.692	&	0.691	&	0.692	\\
+ generated examples: ToxiGen	&	0.658	&	\textbf{0.850}	&	0.742	\\
+ generated examples: Mixed	&	0.664	&	0.778	&	0.716	\\
\hline
\multicolumn{4}{l}{\textbf{SE}} \\		
\hline
No augmentation	&	\textbf{0.642}	&	0.472	&	0.544	\\
+ generated examples: MegaSpeech	&	0.539	&	\textbf{0.743}	&	\textbf{0.625}	\\
+ generated examples: ToxiGen &	0.619	&	0.535	&	0.574	\\
+ generated examples: Mixed	&	0.575	&	0.663	&	0.616	\\
\hline
\multicolumn{4}{l}{\textbf{WS}} \\			\hline		
No augmentation	&	\textbf{0.861}	&	0.673	&	0.756	\\
+ generated examples: MegaSpeech	&	0.853	&	\textbf{0.856}	&	\textbf{0.854}	\\
+ generated examples: ToxiGen	&	0.768	&	0.830	&	0.798	\\
+ generated examples: Mixed	&	0.760	&	0.842	&	0.799	\\
\hline
\multicolumn{4}{l}{\textbf{Average improvement}} \\
\hline
MegaSpeech	& \textbf{-17.8\%}	&	24.3\%	&	\textbf{-0.2\%}	\\
ToxiGen	&	-29.7\%	&	21.4\%	&	-14.0\%	\\
Mixed	&	-30.2\%	&	\textbf{30.53\%}	&	-10.8\%	\\
\hline
\end{tabular}
\captionsetup{font=small,labelfont={bf,sf}}
\caption{Detailed cross-dataset (4-vs-1) results reported in terms of precision, recall and F1 with respect to the hate speech class. For each target dataset, the table repeats the results reported in Table~\ref{tab:cross}, using the RoBERTa-Toxicity model, without and with data augmentation using synthetic examples drawn from our MegaSpeech corpus. In addition, the table shows the results of data augmentation using the same number (240K) of examples drawn from ToxiGen, an alternative corpus of generated hateful and non-hateful text sequences. And, the results of data augmentation where an equal number of examples (120K) is drawn from each source (`Mixed').} 
\label{tab:toxigen}
\end{table}

There are several findings that arise from the results presented in Table~\ref{tab:toxigen}. The bottom part of the table shows the average improvement in precision, recall and F1 across all datasets, having the synthetic examples drawn from either MegaSpeech, ToxiGen, or their balanced mixture. This summary of the results shows similar trends following dataset augmentation for both types of synthetic examples. Specifically, recall rises significantly in both cases, improving by 24.3\% and 21.4\% on average using MegaSpeech and ToxiGen examples, respectively. In both setups, a decrease in precision is incurred due to data augmentation. Overall, there is greater decrease in precision using the ToxiGen (-29.7\%) vs. the MegaSpeech examples (-17.8\%). We conjecture that ToxiGen presents data distribution that is more remotely different from the test set distribution compared with MegaSpeech. Possibly, a larger gap between train and test distributions hurts precision, while the increased diversity among the training examples benefits recall. As in our main results, the factors that affect performance gains (or losses) following data augmentation remain unclear, and are a subject of interest for future inspection and research. 

Another encouraging result that arises from Table~\ref{tab:toxigen} is that the combination of synthetic examples from the two sources yields the highest increase in recall, reaching 30.5\% versus 24.3\% or 21.4\% using the same number of synthetic examples drawn from a single resource. This result supports our conjecture by which increasing train set diversity leads to improved recall in hate speech detection. 

\section{GPT as a Hate Detector}
\label{sec:chatgpt}
\begin{figure}[h]
\centering
\includegraphics[trim=0.5cm 0.5cm 0.5cm 1.5cm, clip,,width=0.7\textwidth]{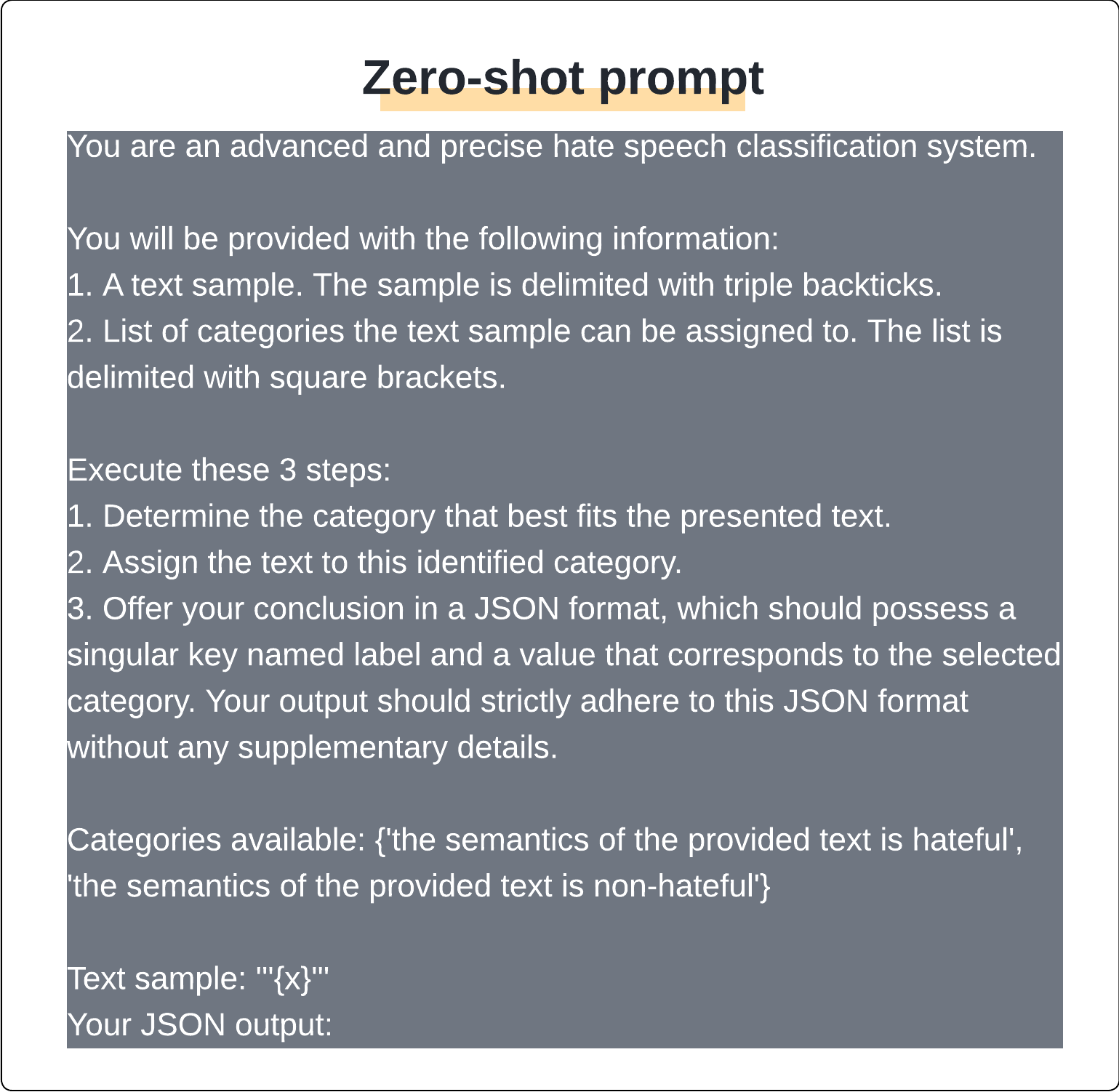}
\captionsetup{font=small,labelfont={bf,sf}}
    \caption{Zero-shot learning prompt of GPT-3.5 (`Text-DaVinci-003').}
    \label{fig:prompt}
\end{figure}

So far, we examined the performance of popular LLMs that have been specialized for hate speech detection. Recently, there have been advances in even larger language models that are pretrained on significantly more text data. These models have also been adapted to generate text that aligns with specific user quality criteria~\cite{ye2023comprehensive}. Remarkably, these models have been shown to perform a variety of text processing tasks, when adequate prompts are provided, even in zero-shot settings, where no explicit examples are provided~\cite{instructgpt22}. Along these lines, researchers have shown that generative LLMs can be used for text labeling, or classification, e.g., when the classification task is phrased as text completion~\cite{chiu2022detecting,gpt3-reduce}. 

Next, we gauge the performance of such a model on our task of hate speech detection, applying a recent GPT-3.5 model, which has been tuned to follow instructions~\cite{instructgpt22}, as a hate detection classifier to our experimental labeled test sets. Importantly, this and similar GPT models are attributed high sensitivity to toxic language\cite{ye2023comprehensive,catgptToxicity23}. It is an eminent question whether such models yield hate speech performance that is nearly perfect--at best, or very high--at least. As detailed below, our experiments show that this is not the case. Furthermore, comparing hate speech performance using a GPT-3.5 model and the other LLMS evaluated in this work, shows that while the GPT model is generally superior, the latter smaller models deliver comparable or better performance in some cases. While we do not evaluate it here, a following research question is whether augmenting the data that very large language models are trained on using data generation at large scale can improve their performance on this task.

\paragraph{Experimental setup.} 

We report our results using 'text-davinci-003', a GPT-3.5 series model~\cite{NEURIPS2020_gpt35},\footnote{https://platform.openai.com/docs/models} which was designed for instruction-following tasks. This choice is motivated by preliminary experiments, in which we manually assessed various recent GPT model variants, and found this model to deliver the most sensible results on a set of reference labeled examples. While other related models may yield different results~\cite{ye2023comprehensive}, systematically optimizing the selection among the existing GPT model variants is beyond the scope of this work.

Figure~\ref{fig:prompt} includes our prompt, phrased in a typical manner to common practice in instructing the model to perform a text classification task~\cite{instructgpt22}. As shown, the prompt first defines the task, instructing the model to perform as a `hate speech classification system'. Then, it details standard procedural text categorization directives, including the requested format of the response. Finally, the target categories are described in natural language: `the semantics of the provided text is hateful', versus, `the semantics of the provided text is non-hateful'. While we experimented with few-shot learning, we found that zero-shot learning, where no examples are provided in addition to this prompt, yielded comparable or better performance. We therefore report our results using a zero-shot classification setting. 

\paragraph{Results.}

\begin{table}[p]
\centering
\begin{tabular}{l|ccc}
    \hline
    & Precision & Recall & F1 \\
\hline
\multicolumn{4}{l}{\textbf{FT}} \\
\hline
Text-davinci	&	0.503	&	\textbf{0.609}	&	\textbf{0.551}	\\
RoBERTa-Tox.	&	0.397	&	0.569	&	0.467	\\
ToxiGen	&	0.439	&	0.421	&	0.430	\\
HateXplain	&	0.471	&	0.500	&	0.485	\\
ToxDect	&	\textbf{0.514}	&	0.445	&	0.477	\\
HateBERT	&	0.373	&	0.536	&	0.440	\\
\hline									
\multicolumn{4}{l}{\textbf{SF}} \\			\hline				
Text-davinci	&	\textbf{0.483}	&	\textbf{0.835}	&	\textbf{0.612}	\\
RoBERTa-Tox.	&	0.468	&	0.742	&	0.574	\\
ToxiGen	&	0.449	&	0.632	&	0.525	\\
HateXplain	&	0.470	&	0.739	&	0.574	\\
ToxDect	&	0.388	&	0.781	&	0.519	\\
HateBERT	&	0.396	&	0.797	&	0.529	\\
\hline
\multicolumn{4}{l}{\textbf{DV}} \\	
\hline
Text-davinci	&	0.609	&	\textbf{0.957}	&	\textbf{0.744}	\\
RoBERTa-Tox.	&	0.692	&	0.691	&	0.692	\\
ToxiGen	&	\textbf{0.698}	&	0.557	&	0.620	\\
HateXplain	&	0.633	&	0.663	&	0.648	\\
ToxDect	&	0.631	&	0.741	&	0.682	\\
HateBERT	&	0.680	&	0.764	&	0.719	\\
\hline
\multicolumn{4}{l}{\textbf{SE}} \\			
\hline
Text-davinci	&	\textbf{0.632}	&	0.745	&	\textbf{0.684}	\\
RoBERTa-Tox.	&	0.539	&	0.743	&	0.625	\\
ToxiGen	&	0.545	&	0.633	&	0.585	\\
HateXplain	&	0.555	&	0.688	&	0.614	\\
ToxDect	&	0.517	&	0.739	&	0.608	\\
HateBERT	&	0.539	&	\textbf{0.764}	&	0.632	\\
\hline
\multicolumn{4}{l}{\textbf{WS}} \\			\hline		
Text-davinci	&	0.690	&	0.845	&	0.761	\\
RoBERTa-Tox.	&	0.853	&	0.856	&	0.854	\\
ToxiGen	&	\textbf{0.881}	&	0.790	&	0.833	\\
HateXplain	&	0.848	&	0.865	&	0.856	\\
ToxDect	&	0.831	&	0.902	&	\textbf{0.865}	\\
HateBERT	&	0.753	&	\textbf{0.919}	&	0.828	\\
\hline
\end{tabular}
\captionsetup{font=small,labelfont={bf,sf}} 
\caption{Cross-dataset (4-vs-1) results reported in terms of precision, recall and F1 with respect to the hate speech class. For each target dataset, the table repeats the results using the various LLMs which we finetuned using the other four labeled datasets, augmented with 240K synthetic examples drawn from the MegaSpeech corpus as  reported in Table~\ref{tab:cross}. The top line of results for each of the target datasets show the evaluation of Text DaVinci, a recent GPT-3.5 model that has been adapted to follow instructions in natural language. The prompt used to elicit class labels using Text Davinci is shown in Figure~\ref{fig:prompt}.} 
\label{tab:chatgpt}
\end{table}

Table~\ref{tab:chatgpt} shows hate speech detection results using the GPT-3.5 model. In order to minimize computational costs, we applied and report the performance of the model on 20\% randomly selected examples of each test set. 

It is apparent from the table that this model has been trained to be sensitive to toxicity. In each of the individual experiments, recall performance of this model is higher than its precision performance. Overall, across the experiments, recall ranges between 0.61-0.96, whereas precision ranges between 0.48-0.69. We find that this affirms the high importance that is attributed to detecting text sequences that are potentially hateful. 

In order to allow direct comparison with the other LLMs evaluated in this work, Table~\ref{tab:chatgpt} repeats relevant results using the various LLMs that are included in Table~\ref{tab:cross}. Maintaining our focus on recall performance, we consider the models that were finetuned using the synthetic examples, as data augmentation achieves higher recall rates. While the DaVinchi model is evaluated on 20\% of the examples, and the other models are evaluated on the full test sets, we believe that the comparison shows general trends of interest. As shown, the GPT-3.5 model achieves the best performance in most of the experiments. Yet, it is striking that the smaller LLMs achieve higher performance in some cases. Specifically, HateBERT, a variant of BERT, which is smaller by magnitudes of order from GPT-3.5, yields the best recall performance in 2/5 of the experiments. (Interestingly, these are the two experiments in which the largest improvements were obtained following data augmentation, as reported in Table~\ref{tab:cross}.) 

Overall, the following insights arise from Table~\ref{tab:chatgpt}. First, it shows that state-of-the-art LLMs, which show unprecedented capabilities of natural language understanding, still struggle in identifying hate as intended or perceived in human-authored texts. Thus, despite recent advancements, hate speech detection remains an open pending problem. Second, we find it encouraging that relatively small models sometimes yield hate speech detection performance that surpasses models like GPT-3.5. Considering the improvements in recall performance that we observed following data augmentation in finetuning these models, we believe that further scaling the data that models like the recent variants of GPT are trained on by means of targeted text generation holds promise for improving state-of-the-art performance on this task. 

\section{Conclusion}
\label{sec:conclusion}
This work evaluated several large transformer-based language models that have been specialized for the task of hate speech detection using task-related data. The key findings are that augmenting training data for those hate speech classifiers with generated examples leads to substantial gains in recall, at the cost of some precision loss. This indicates that the synthetic data introduces useful lexical diversity, while potentially adding some noise. Overall F1 scores still tend to improve. Among the models tested, the large generative model GPT-3.5 showed strong hate speech detection capabilities even in a zero-shot setting, though smaller specialized models like HateBERT were competitive in some cases. As future research, we believe that extending and enhancing data generation efforts, using state-of-the-art generative LLMs, may yield even higher quality synthetic text sequences. Scaling up these efforts may also support enhanced training of very large language models, to various semantic phenomena, including hate speech.
\newpage

\bibliography{final}
\bibliographystyle{apalike}

\end{document}